\documentclass[preprint,12pt]{elsarticle}

\usepackage{amssymb}
\usepackage{amsthm}
\usepackage{amsmath}
\usepackage{caption}
\usepackage{subcaption}
\usepackage{color}
\usepackage{comment}
\usepackage{hyperref}
\usepackage{xcolor}

\journal{Chemometrics and Intelligent Laboratory Systems}

\newcommand{\m}[1]{\boldsymbol{#1}}
\let\m\mathbf
\newcommand{\T }{^{\text{T}}}
\newcommand{\norm}[1]{\left\lVert#1\right\rVert}
\newcommand{\rnl}[1]{\textcolor{red}{\textbf{[RNL]} #1}}
\newcommand{\rjr}[1]{\textcolor{purple}{\textbf{[RJR]} #1}}

\let\m\mathbf

\begin{document}

\begin{frontmatter}

%% Title, authors and addresses

%% use the tnoteref command within \title for footnotes;
%% use the tnotetext command for theassociated footnote;
%% use the fnref command within \author or \address for footnotes;
%% use the fntext command for theassociated footnote;
%% use the corref command within \author for corresponding author footnotes;
%% use the cortext command for theassociated footnote;
%% use the ead command for the email address,
%% and the form \ead[url] for the home page:
%% \title{Title\tnoteref{label1}}
%% \tnotetext[label1]{}
%% \author{Name\corref{cor1}\fnref{label2}}
%% \ead{email address}
%% \ead[url]{home page}
%% \fntext[label2]{}
%% \cortext[cor1]{}
%% \affiliation{organization={},
%%             addressline={},
%%             city={},
%%             postcode={},
%%             state={},
%%             country={}}
%% \fntext[label3]{}

\title{Opening the black-box of Neighbor Embedding with Hotelling's $T^2$ statistic and $Q$-residuals}
%% TODO: Evtl sowas wie "Interpretability of Dimensionality Reduction Dimensions" ???

%% use optional labels to link authors explicitly to addresses:
%% \author[label1,label2]{}
%% \affiliation[label1]{organization={},
%%             addressline={},
%%             city={},
%%             postcode={},
%%             state={},
%%             country={}}
%%
%% \affiliation[label2]{organization={},
%%             addressline={},
%%             city={},
%%             postcode={},
%%             state={},
%%             country={}}

\author[inst1]{Roman Josef Rainer}

\affiliation[inst1]{organization={Software Competence Center Hagenberg},%Department and Organization
            addressline={Softwarepark 32a}, 
            city={Hagenberg},
            postcode={4232}, 
            state={},
            country={Austria}}
\author[inst1]{Michael Mayr} 
\author[inst1]{Johannes Himmelbauer}
\author[inst1]{Ramin Nikzad-Langerodi}

%\affiliation[inst2]{organization={Department Two},%Department and Organization
%            addressline={Address Two}, 
%            city={City Two},
%            postcode={22222}, 
%            state={State Two},
%            country={Country Two}}

\begin{abstract}
%% Text of abstract
In contrast to classical techniques for exploratory analysis of high-dimensional data sets, such as principal component analysis (PCA), neighbor embedding (NE) techniques tend to better preserve the local structure/topology of high-dimensional data. However, the ability to preserve local structure comes at the expense of interpretability: Techniques such as t-Distributed Stochastic Neighbor Embedding (t-SNE) or Uniform Manifold Approximation and Projection (UMAP) do not give insights into which input variables underlie the topological (cluster) structure seen in the corresponding embedding. We here propose different "tricks" from the chemometrics field based on PCA, $Q$-residuals and Hotelling's $T^2$ contributions in combination with novel visualization approaches to derive local and global explanations of neighbor embedding. We show how our approach is capable of identifying discriminatory features between groups of data points that remain unnoticed when exploring NEs using standard univariate or multivariate approaches.
\end{abstract}

%%Graphical abstract
%\begin{graphicalabstract}
%\includegraphics{grabs}
%\end{graphicalabstract}

%%Research highlights
%\begin{highlights}
%\item Research highlight 1
%\item Research highlight 2
%\end{highlights}

\begin{keyword}
%% keywords here, in the form: keyword \sep keyword
neighbor embedding \sep t-SNE \sep UMAP \sep PCA \sep chemometrics
%% PACS codes here, in the form: \PACS code \sep code
%\PACS 0000 \sep 1111
%% MSC codes here, in the form: \MSC code \sep code
%% or \MSC[2008] code \sep code (2000 is the default)
%\MSC 0000 \sep 1111
\end{keyword}

\end{frontmatter}

%% TODO: was macht das? funktioniert nicht, compiler error
%\linenumbers

%% main text
\section{Introduction}
\label{sec:introduction}
Neighbor embedding techniques have gained increasing popularity in a wide variety of scientific disciplines such as machine learning \cite{van2008visualizing}, biology \cite{li2017application}, physics \cite{chang2004super} or engineering \cite{hajibabaee2021empirical}. In contrast to classical techniques for exploratory analysis of high-dimensional data sets, such as principal component analysis (PCA), these methods tend to better preserve the "intrinsic structure" or (local) topology of high-dimensional data when mapped to (i.e. embedded in) low-dimensional spaces which often provides additional insights. However, the ability to preserve local structure comes at the expense of interpretability: Techniques such as t-Distributed Stochastic Neighbor Embedding (t-SNE) \cite{van2008visualizing} or Uniform Manifold Approximation and Projection (UMAP) \cite{mcinnes2018umap} do not give insights into which input variables underlie the topological (cluster) structure seen in the corresponding embedding. Tools, such as Embedding Projector (EP) \cite{smilkov2016embedding} or t-viSNE \cite{chatzimparmpas2020t} can help to better understand the corresponding embedding. EP provides a web-based, interactive environment that allows different "views" on high-dimensional data sets and their embedding. However, while EP is useful to explore  embedding in terms of pre-defined labels, there is no possibility to investigate the relationship between (semantically) meaningful directions in the embedded space and the input variables. In general, inspection of how single variables change across an embedding is straightforward and can be undertaken for low-dimensional data sets but becomes prohibitive for data sets with hundreds of thousands of variables. In order to address this issue, t-viSNE comes with a so-called "dimension correlation tool", which allows to explore the correlation between the input variables and a user-defined (e.g. local and/or non-linear) direction in the embedded space. Bibal et al. (\cite{bibal2020explaining}) employ a modified LIME (local interpretable model-agnostic explanations) approach  to explain t-SNE embedding. Similar to t-viSNE, this approach derives "explanations" for local neighborhoods around particular data points in terms of the input variables. Both methods derive local explanations rather than explaining the (global) directions of an embedding and require considerable computational resources for data sampling and local modelling. In addition, both methods require considerable user-interactions, which might restrict their use to expert users familiar with the techniques. An alternative approach has been recently proposed in \cite{andries2022dual}, where the authors use local neighborhood information derived from UMAP as an additional constraint in PCA. However, for some data sets, replication of the (non-linear) neighborhood structure with a linear model is difficult and eventually yields latent variables (LVs) that capture the noise rather than the systematic variation in the data.

\begin{figure}
    \centering
    \includegraphics[width=0.95\columnwidth]{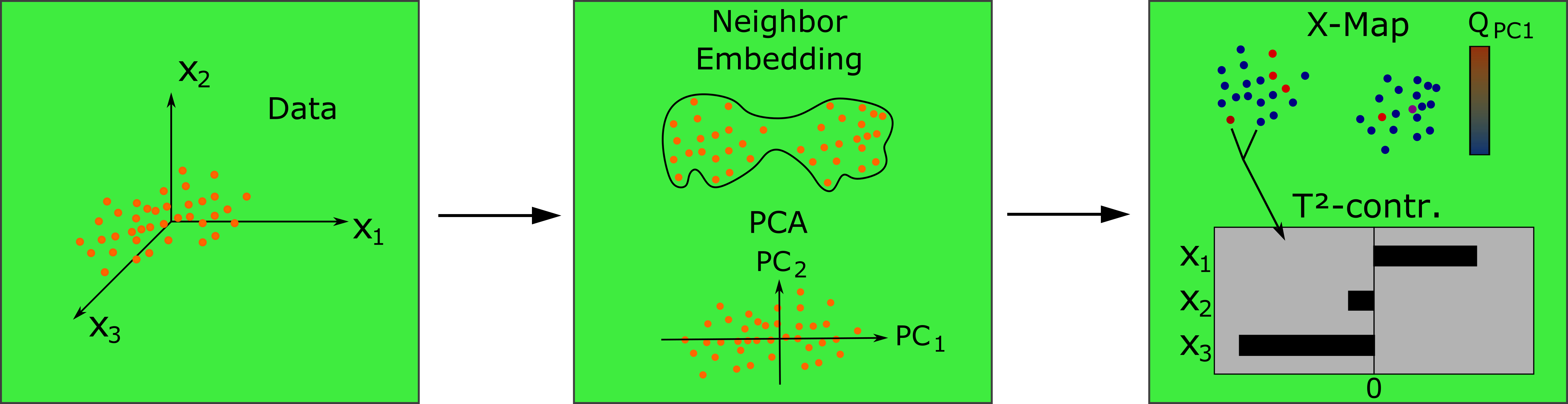}
    \caption{PCA-derived explanations of neighbor embedding by means of relative Hotelling's $T^2$ contributions and $Q$-residuals.}
    \label{fig:our_approach}
\end{figure}

In the current contribution, we propose a simple, yet effective, workflow to derive explanations of local and global directions of (non-linear) embedding that is computationally efficient and reduces the overhead of user-interactions. In brief, we first create the corresponding embedding and fit a PCA model to the input data. We then employ relative Hotelling's $T^2$-contributions to derive explanations for the difference between individual data points or entire clusters. In addition, we propose novel visualizations that include the PCs scores along with the corresponding $Q$-residuals in order to understand which data points are well characterized by individual LVs underlying the data set. The main components of our approach are summarized in Figure \ref{fig:our_approach}. % minor grammar corrections

We will focus on deriving PCA-based explanations of UMAP embedding, which build upon t-SNE and now belongs to the most widely used neighbor embedding methods in the data science community. In general, however, the same workflow can be applied to any (non-linear/parametric) dimensionality reduction method. We refer to the excellent review by Wang et al. \cite{wang2020understanding} for an overview of the current state-of-the-art dimension reduction techniques. % minor grammar corrections 

\section{Theory}
\label{sec:theory}

\subsection{Principal Component Analysis (PCA)}
\label{sub:linear_projections}
Principal component analysis (PCA) is among the most widely used techniques for exploratory analysis of multivariate datasets \cite{jolliffe2005principal}. PCA decomposes an $I\times J$ (e.g. time points $\times$ process variables) matrix $\m{X}$ into a set of \emph{scores} $\m{T} = [\m t_1,\dots, \m t_A]^{I\times A}$ and \emph{loadings} $\m{P}= [\m p_1,\dots, \m p_A]^{J\times A}$ vectors such that \cite{wold1987principal}
\begin{align}
\label{eq:pca}
\begin{aligned}
& \m X  = \sum_{i=1}^A \m t_i\m p_i^T + \m E.\\
& \textbf{s.t.} \;\; \m t_i^T\m t_j = 0 \; \text{and} \; \m p_i^T\m p_j = 0 \;\; \text{for} \;\; i \neq j.
\end{aligned}
\end{align}
Ideally, $A$ in Eq.\eqref{eq:pca} is chosen such that the systematic variation in $\m X$ is encoded in the mutually orthogonal principal components (PCs) $\m t_1\m p_1^T, \dots, \m t_A\m p_A^T$ while the random noise is encoded in the $I\times J$ residual matrix $\m E$. The strength of PCA is rooted in the fact that each PC represents a separate source of variation underlying the observed data that can be analyzed at the samples (scores) and variables (loadings) level. While the former allows to detect (e.g. temporal) trends in (process) data, the latter allows to investigate the correlations between the original variables and those trends. In addition, computation of the variance explained by each PC is straightforward, allowing to quantify the contribution of a trend to the total variance in a data set. Finally, PCA fits a bilinear model to the data, which can be useful when aiming at making predictions on new data. This is exploited e.g. in multivariate statistical process control (MSPC) to detect departure from normal operating conditions (\cite{ferrer2007multivariate}) or in classification problems (\cite{wold1977simca}).   % maybe smth. like : In addition, computing the explained variance by each PC is straightforward, quantifying a trend's contribution to the total variance in a data set.

Despite its versatility, several characteristics of PCA are less favorable for data analysis. First, it is important to note that the objective of PCA is to find directions in the high-dimensional space of the input variables where the (explained) variance is largest. Consequently, PCA is good at capturing global trends while largely neglecting local structures that might be present and eventually encode important information about the process under investigation. On the other hand, the latent variables (LVs) that span the PCA subspace are linear combinations of the original variables. According to the central limit theorem, linear combinations of random variables converge to normal distributed LVs \cite{nikzad2020domain} as the number of variables increases and scores plots thus often tend to show spherically shaped clusters where the natural topology of the data is "distorted" which is particularly the case for very high-dimensional data

Over the past decades, several dimension reduction methods have been proposed that better preserve the (local) structure/topology of high dimensional data, two of which we find particularly useful. These shall thus be introduced in the next sections. These will therefore be briefly introduced in the following sections

%\paragraph{Laplacian-based Methods}
%{\bf Ramin}
%
%\begin{itemize}
%    \item Laplacian Eigenmaps
%    \item Locality preserving projections
%    \item ...
%\end{itemize}

\subsection{Neighbor Embedding}
\label{sub:nonlinear_projections}
The general idea of Neighbor Embedding (NE) is to preserve (local) distances between data points in the high-dimensional space when deriving the corresponding low-dimensional representation.  

\paragraph{t-SNE} t-SNE models the distance between two data points $\m x_i$ and $\m x_j$ in the high-dimensional space as the (conditional) probability 

\begin{equation}
p_{ij} = \frac{p_{j|i} + p_{i|j}}{2N},
\label{eq:tsne1}
\end{equation}
with
\begin{equation}
\begin{aligned}
p_{j|i} = \frac{v_{j|i}}{\sum_{k \neq i}v_{j|i}} &; &   v_{j|i} = \text{exp}\left(-\frac{\norm{\m x_i - \m x_j}^2_2}{2\sigma_i^2}\right).
\label{eq:tsne2}
\end{aligned}
\end{equation}
$N$ denotes the number of data points, and $\sigma_i$ (i.e. the Gaussian kernel bandwidth) controls how fast the probability decreases with increasing Euclidean distance from the $i$-th data point. Instead of manually choosing the (optimal) bandwidth for each data point, t-SNE "automatically" adjusts $\sigma$ to be small in dense regions and large in sparse regions of the input space based on the so-called perplexity parameter
\begin{equation}
\text{Perplexity}(P_i) = 2^{-\sum p_{j|i} \log_2 p_{j|i}}
\label{eq:tsne_perplexity}
\end{equation}
that needs to be specified by the user. The perplexity is related to the number of (close) neighbors around each data point that should be considered in the embedding. A large perplexity emphasizes global while a small one focuses on local structure. The distance in the (low-dimensional) embedded space is modeled using Student's t-distribution with the corresponding probability
\begin{equation}
\begin{aligned}
q_{ij} = \frac{w_{ij}}{\sum_{k \neq l}w_{kl}}  &; & w_{ij} = \frac{1}{1 + \norm{\m y_i - \m y_j}^2_2},
\label{eq:tsne3}
\end{aligned}
\end{equation}
where $\m y_i$ denotes the (two-dimensional) coordinates of the $i$-th data point in the embedded space. Finally, the so-called \emph{Kullback-Leibler} (KL) divergence between the high- and low dimensional distributions $P$ and $Q$ with respect to these coordinates is minimized, i.e.   

\begin{equation}
\min_{\m y_i, \dots, \m y_N} \; \text{KL}(P|| Q)
\label{eq:tsne_cost}
\end{equation}
with $\text{KL}(P|| Q) = \sum_{i \neq j} p_{ij} \log \frac{p_{ij}}{q_{ij}}$. Technically, this is achieved by means of gradient descent using some random initialization of the $y$-coordinates.

\paragraph{UMAP} Conceptually, t-SNE and UMAP share the same basic idea with subtle differences related to how similarity is modelled. Most importantly, UMAP does not normalize pair-wise distances, i.e. $\sum_{i,j} p_{ij} \neq 1$ (the same is true for $q_{ij}$), employs
\begin{equation}
p_{j|i} = \text{exp}\left(-\frac{\text{d}(\m x_i, \m x_j) - \rho}{\sigma}\right)
\label{eq:umap_sim}
\end{equation}
as distance metric and minimizes cross-entropy
\begin{equation}
 \text{CE}(P||Q) = \sum_{i\neq j} \left( p_{ij} \cdot \log \left(\frac{p_{ij}}{q_{ij}}\right) + (1 - p_{ij}) \cdot \log \left(\frac{1-p_{ij}}{1 - q_{ij}}\right)\right)
 \label{eq:cross_entropy}
\end{equation}
instead of KL-divergence. The parameter $\rho$ in Eq.\eqref{eq:umap_sim} represents the distance from the $i$-th data point to its nearest neighbor and implies a different distance metric for each data point (that in turn requires a symetrization that is slightly different from Eq. \eqref{eq:tsne1}). $\text{d}(\m x_i, \m x_j)$ denotes some distance (e.g. Euclidean distance) between $\m x_i$ and $\m x_j$. In addition, UMAP uses the number of $k$ nearest neighbors
\begin{equation}
k = 2^{-\sum p_{ij}}
\label{eq:umap_neighbor}
\end{equation}
when computing $p_{j|i}$ and the family of curves
\begin{equation}
q_{ij} = \left(1 + a \cdot (\norm{\m y_i-\m y_j}^2_2)^b\right)^{-1}
\label{eq:umap_sim_y}
\end{equation}
instead of perplexity and Student t-distribution, respectively. For the latter, the parameters $a$ and $b$ are obtained by non-linear least-squares fitting.

\paragraph{t-SNE vs. UMAP} The fact, that UMAP does not normalize the "probabilities" not only makes it faster (due to omission of sum computations) and consume less memory (replacement of gradient descent by stochastic gradient descent) but also allows one to "map" new samples to the embedding without changing the position of the training samples, e.g. by computing
\begin{equation}
\hat{\m y}_{N+1} = \min_{\m y_{N+1}} \; \text{CE}(P_{\m y_1,\dots,\m y_{N+1}}||Q_{\m y_1,\dots,\m y_{N+1}}).
\label{eq:umap_projection}
\end{equation}  
With t-SNE this is not possible, since any new data point changes $p_{j|i}$ and $q_{j|i}$ (through normalization) and thus the (optimal) coordinates of the training samples. In addition, UMAP better balances global and local structure preservation. This is because data points that are far apart (small $p_{ij}$) contribute little to the KL-divergence but contribute to CE through the second term in Eq. \eqref{eq:cross_entropy}. 

\begin{comment}
In general, UMAP scales better \rnl{to large datasets? higher dimensions?}, is faster, consumes less memory \rnl{why?} than t-SNE. UMAP preserves global structures better,
and is at least compatitive in local structures to t-SNE. This is achieved by doing no normalization, defining the
number of neighbors without log-function, the choice of cost function, and using stochastic gradient descent instead of
gradient descent.
\end{comment}

\subsection{Interpretability}
\label{sub: interpretability}
Unlike in PCA, where each embedded sample can be expressed as linear transformation of the corresponding inputs, i.e.
\begin{equation}
\m y_i = \m x_i\T\m P,
\label{eq:pca_proj}
\end{equation}
the (non-linear) map $f:\mathcal{X}\rightarrow\mathcal{Y}$ remains a black-box in both t-SNE and UMAP. Since there is no model, interpretability is not given.
\begin{comment}
\rnl{this should be discussed in more detail including previous studies that have addressed this issue (from introduction)}
A major issue of dimensionality reduction methods like t-SNE and UMAP is the interpretability of the embedded
space. Even though data can afterwards be projected into the embedded space with UMAP \rnl{how?! Why can it be not done with t-SNE?}, the dimensions of the
embedded space have no meaning, unlike in PCA or other linear methods. \rnl{input Ramin}. 
\end{comment}
To mitigate this issue we propose two
complementary workflows to better understand the embedded space.

\subsection{Our approach}
\label{sub:our}
The idea of using PCA as preprocessing step or to use PCA scores (i.e. individuals columns of the $\m T$ matrix from Eq. \eqref{eq:pca}) as color code for NE has been proposed earlier (e.g in \cite{chatzimparmpas2020t}) and is a useful tool to get a quick overview about which variables contribute to the topological structure of the embedding. We extend this idea by introducing two concepts widely used in chemometrics: $Q$-residuals and relative $T^2$-contributions, and introduce Voronoi plots for visualization of the corresponding NE. 

\begin{figure}
    \centering
    \includegraphics{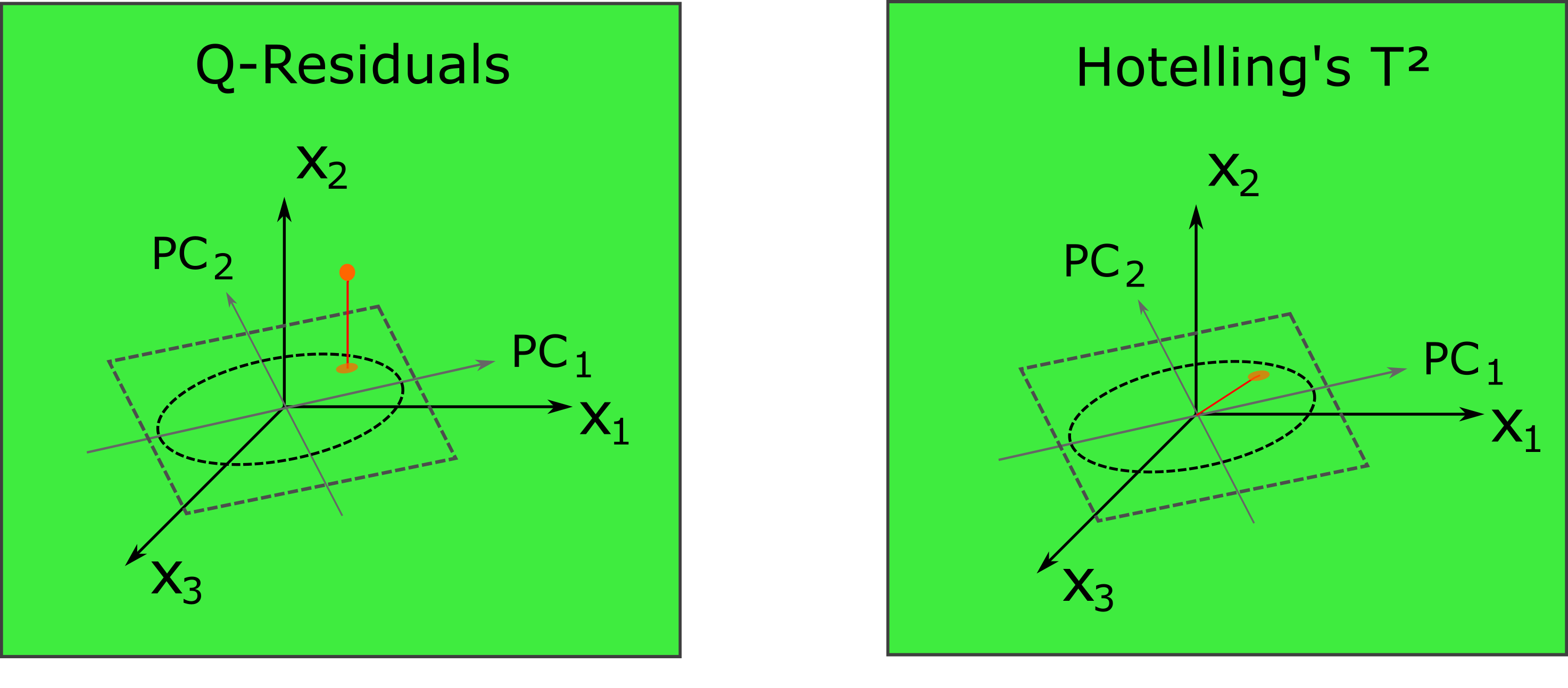}
    \caption[width=0.75\columnwidth]{$Q$-Residuals and Hotelling's $T^2$-statistic.}
    \label{fig:T2_Q}
\end{figure}

\paragraph{Q-Residuals} are also known as reconstruction loss in the machine learning community. For an input sample $\m x_i$ 
\begin{equation}
Q_i = \m x_i(\m I - \m P_A\T\m P_A)\m x_i\T,
\label{eq:qres}
\end{equation} 
with $\m P_A$ denoting the loadings matrix of a PCA model with $A$ components. Samples with low $Q$-residuals are well represented by the model, i.e. they have small (orthogonal) distances to their low-dimensional projections (Figure \ref{fig:T2_Q}). When using the PC scores $\m t_k$ for $k\in {1,\dots, A}$ as color code for the embedded data, we here propose to always also display the corresponding $Q$-residuals in order to see for which samples the corresponding "explanations" have strong (small $Q$) and weak (large $Q$) support. For the latter, a different PC might better explain their location in the embedding. PC specific $Q$-residuals are calculated by replacing $\m P_A$ in Eq.\eqref{eq:qres} by the $k$-th loadings vector $\m p_k$.  

\begin{figure}[t!]
  \centering
		\includegraphics[width=0.99\columnwidth]{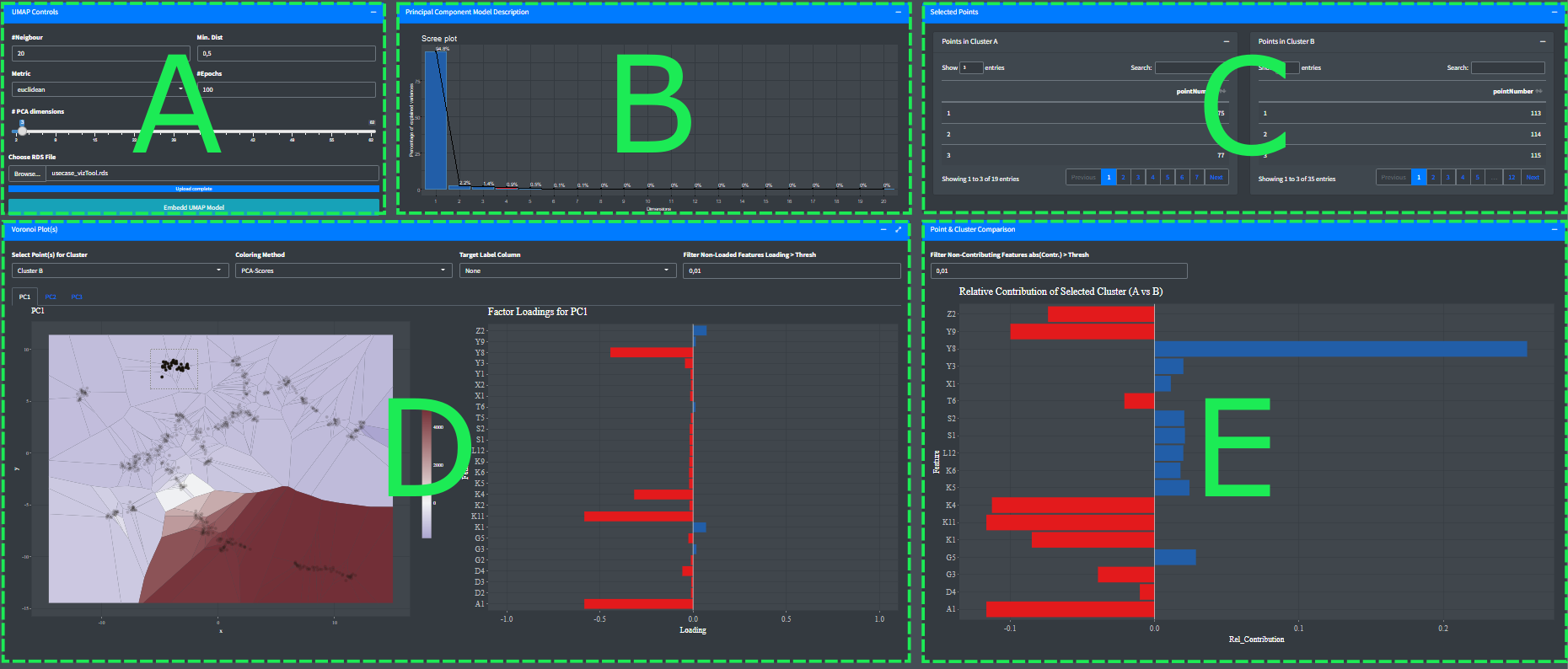}
  \caption{The X-MAP tool at a glance. A) Data import, UMAP settings and number of PC's. B) Explained variance per PC. C) Selected data points from the UMAP embedding. D) Voronoi plot of the embedded data (color-coded 
  either by PCA score, $Q$-residuals or a single input variable) and loadings of the corresponding PC. E) Relative $T^2$ contributions explaining the difference between two data points or clusters selected in D.}
  \label{fig:viztool_overview}
\end{figure}

\paragraph{Relative T$^2$-contributions} The Hotelling's $T^2$-statistic is a measure of outlingness w.r.t. the centroid of the (training) data within the PC space (Figure \ref{fig:T2_Q}). The corresponding $T^2$-contributions are a measure of influence of each input variable to the outlingness. For the $i$-th sample
\begin{equation}
\m t_{\text{cont},i} = \m t_i\m\Lambda^{-1/2}\m P_A,
\label{eq:Tcon}
\end{equation}    
with $\m \Lambda$ being a diagonal matrix holding the $A$ leading Eigenvalues of $\m X\T\m X$\footnote{\url{https://wiki.eigenvector.com/index.php?title=T-Squared_Q_residuals_and_Contributions}}. The corresponding scaling takes care that the same weight is assigned to the contribution from each subspace dimension to the explanations. Relative $T^2$-contributions on the other hand, can be used to investigate which variables contribute most to the difference between $\m x_i$ and $\m x_j$ w.r.t. the ($A$-dimensional) PC space:
\begin{equation}
\m t_{\text{cont},ij} = \m t_{\text{cont},i} - \m t_{\text{cont},j}
\label{eq:rel_Tcon}
\end{equation} 
The same approach can be used to derive explanations for why entire clusters within an embedding differ from each other. To do so, the contributions are computed with respect to the corresponding cluster centroids.   

\paragraph{Voronoi plots}
In order to better explore the non-linear structure of the embedded space, we propose Voronoi diagrams. A Voronoi diagram is a partitioning of a plane into regions with each point in that region being closest to a single data point in terms of Euclidean distance. This allows to see the value of a statistic to be displayed (e.g. $Q$-Residuals) for individual data points over the whole embedded space.

\paragraph{Implementation} A prototypical implementation of the entire data exploration workflow including graphical user interface (GUI) was undertaken using the shiny web application framework for R \citep{chang2017shiny}. We provide open source code at https://github.com/RNL1/XMAP.git.

\begin{figure}[t!]
  \centering
		\includegraphics[width=0.99\columnwidth]{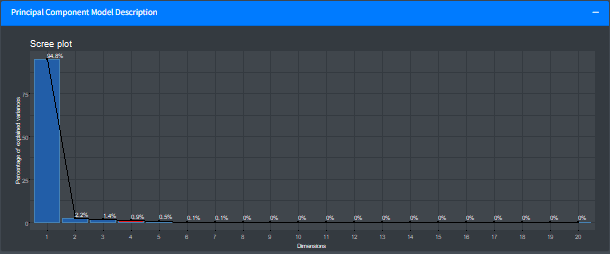}
  \caption{Fraction of explained variance for each PC. The number of PCs included in the model is indicated by the red bar (\#PCs = 4).}
  \label{fig:viztool_pca}
\end{figure}

\section{X-Map GUI}
\label{sec:Viztool}
Figure \ref{fig:viztool_overview} outlines the functionality of the proposed data exploration tool that we coin explainable map (X-MAP). First, a .rds file holding the data to be explored must be provided in long format followed by specifying the UMAP parameters, i.e. number of nearest neighbors, minimum distance (see Eq. \eqref{eq:umap_sim}, similarity metric, and number of training epochs (Figure \ref{fig:viztool_overview}A). Subsequently, the maximum number of PC's is specified followed by fitting the PCA model and UMAP embedding. Once model fitting has completed, the percentage of variance explained by each PC appears in a separate subplot (Figure \ref{fig:viztool_pca}). 

In the example shown, more than 94\% of the variance is explained by the first PC. Note that an optimal number of PCs to be retained is proposed and automatically selected during model fitting but can be manually adjusted after completion of the training epochs. 

\begin{figure}[t!]
  \centering
		\includegraphics[width=0.99\columnwidth]{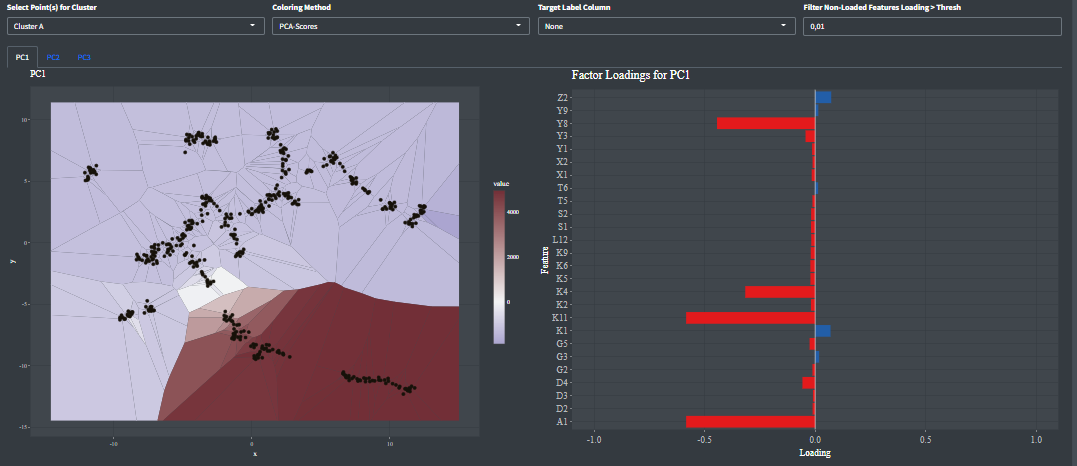}
  \caption{Voronoi diagram color-coded according to PC1 scores (left) and corresponding loadings (right).}
  \label{fig:viztool_pc1}
\end{figure}

In addition to the explained variance plot, a Voronoi diagram of the UMAP embedding appears with the polygones being by default colored according to the score of PC 1 side by side with the corresponding loadings (Figure \ref{fig:viztool_pc1}).

\begin{figure}[h!]
  \centering
		\includegraphics[height = 6.5cm]{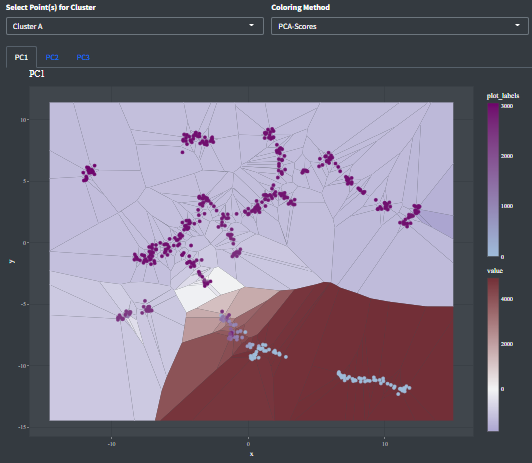}
  \caption{Data points colored according to Y8 variable.}
  \label{fig:viztool_Y8}
\end{figure}

\begin{figure}[h!]
  \centering
		\includegraphics[width=0.99\columnwidth]{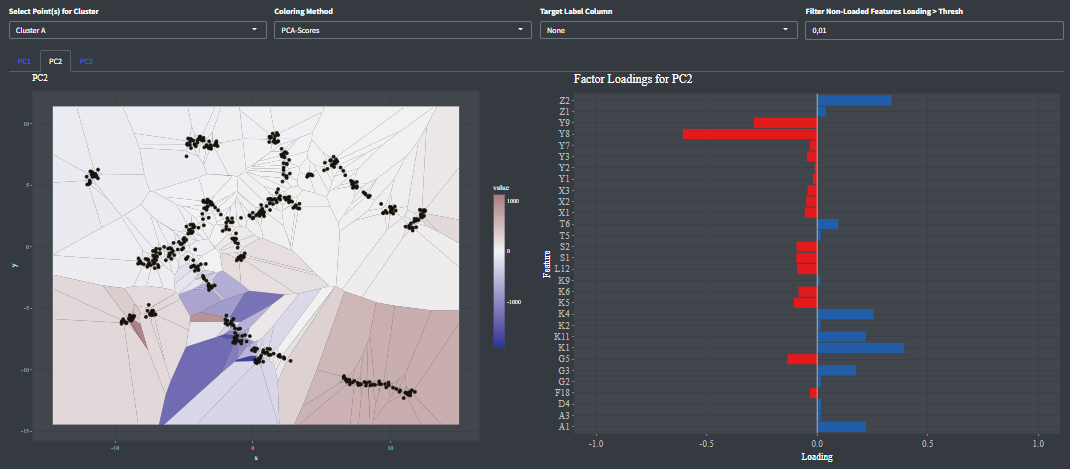}
  \caption{Voronoi diagram color-coded according to PC2 scores (left) and corresponding loadings (right).}
  \label{fig:viztool_pc2}
\end{figure}

\begin{figure}[tbp]
\centering
  \begin{subfigure}[b]{0.48\columnwidth}
\centering
		\includegraphics[width=0.99\columnwidth]{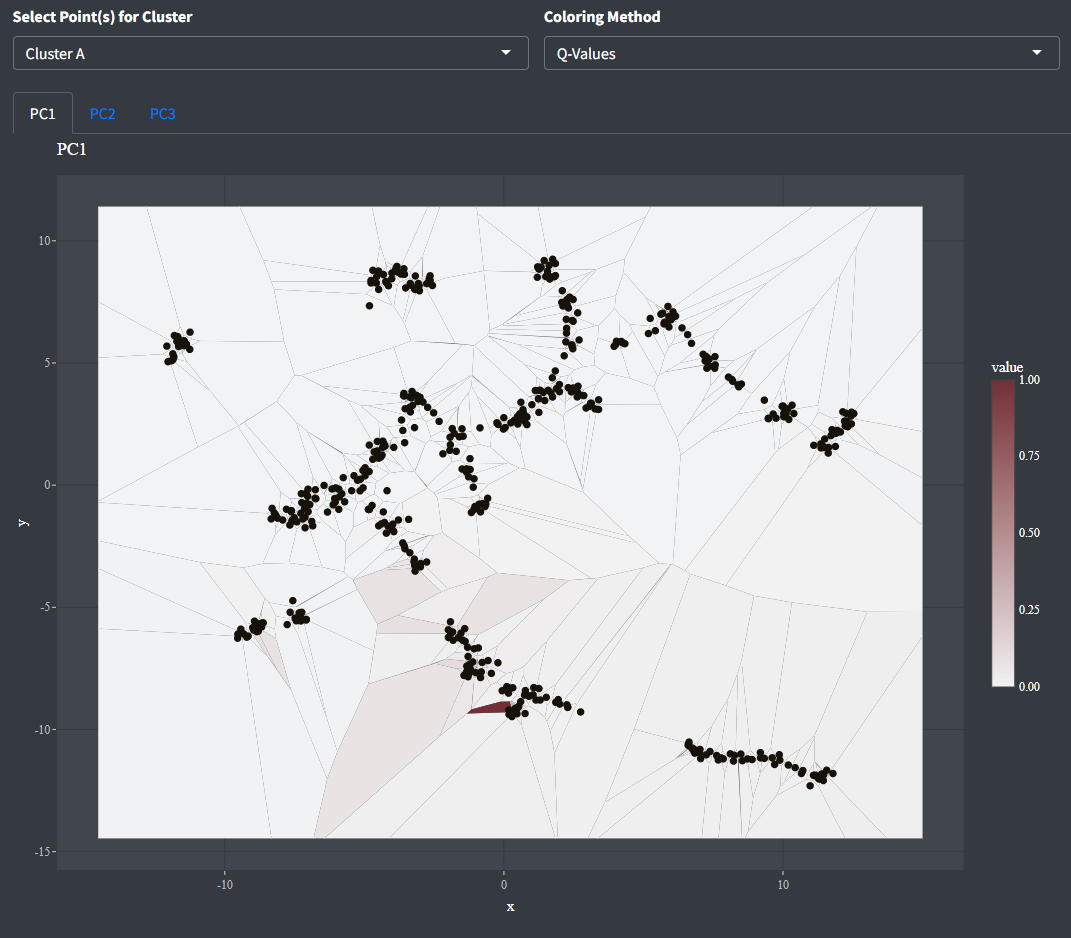}
  \captionof{figure}{Principal component 1}
  \label{fig:viztool_qval}
  \end{subfigure}
  \hfill
  \begin{subfigure}[b]{0.48\columnwidth}
  \centering
		\includegraphics[width=0.99\columnwidth]{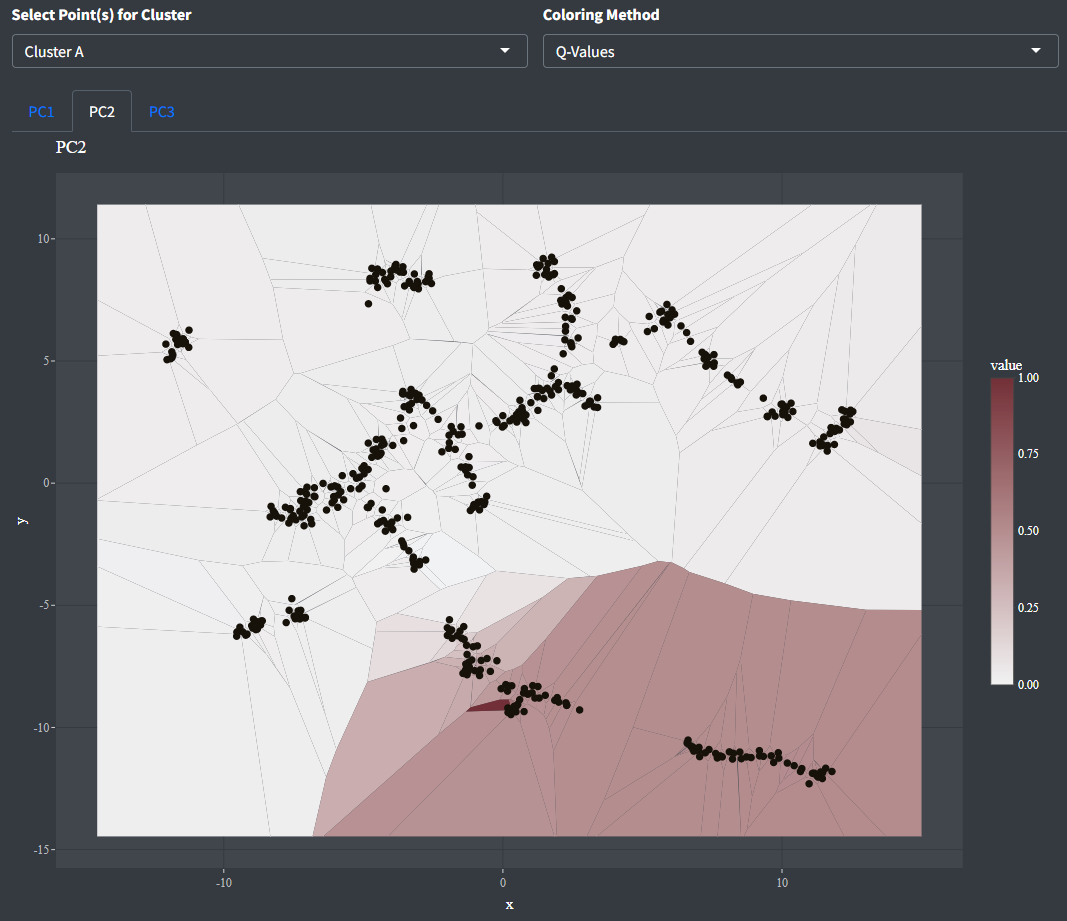}
  \captionof{figure}{Principal component 2}
  \label{fig:viztool_qval2}
  \end{subfigure}
  \caption{PC-specific $Q$-Residuals. Note that $Q$-values are normalized between 0 and 1.}
  \label{fig:viztool_qvals}
\end{figure}

In the example shown, data points located in the lower right corner of the embedding exhibit a higher score on PC 1 compared to the points located in the blue areas. Separation between the two areas is mostly due to lower values of variables Y8, K4, K11 and A1 for the former as can be seen by inspecting the corresponding loadings. Figure \ref{fig:viztool_Y8} shows the same embedding with the data points color-coded according to the value of variable Y8, which is in-line with this interpretation. 
PC 2, on the other hand, encodes information about differences between clusters in the lower half of the UMAP (Figure \ref{fig:viztool_pc2}).
An important criterion for PCA-based interpretation of a non-linear embedding like UMAP is, how well each data point is represented by the current PC. In the worst case, a PC that explains a high proportion of variance in the data (i.e. PC 1 in our example) only explains a small portion of the information contained in some data points. $Q$-Residuals are a means of diagnosing if some data points, i.e. outliers, are poorly represented. Figure \ref{fig:viztool_qvals}a and \ref{fig:viztool_qvals}b  show the same Voronoi diagramm of the embedding with the polygones color-coded according to the $Q$-residuals with respect to the first and second PC. Note that the values are min-max normalized and lie between 0 and 1 for samples with the lowest and highest $Q$-residuum in the data set, respectively. Notably, a single data point appears as potential outlier (with respect to the first and second PC) and thus warrants further investigation. For PC~2, $Q$-residuals indicate that samples located in the lower right corner are poorly represented. Thus, the corresponding loadings might provide limited information about these samples.         

\begin{figure}[tbp]
\centering
  \begin{subfigure}[b]{0.48\columnwidth}
\centering
		\includegraphics[width=0.99\columnwidth]{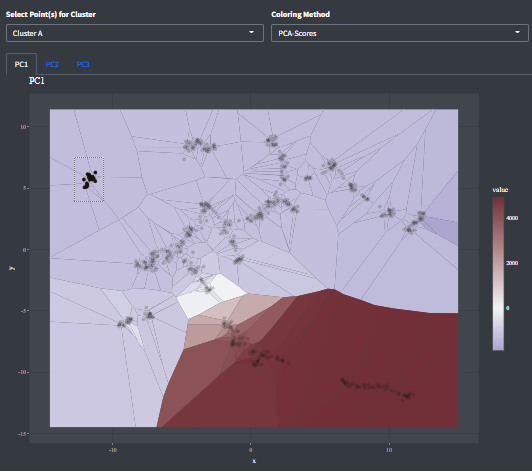}
  \captionof{figure}{Cluster A}
  \label{fig:viztool_clusterA}
  \end{subfigure}
  \hfill
  \begin{subfigure}[b]{0.48\columnwidth}
  \centering
		\includegraphics[width=0.99\columnwidth]{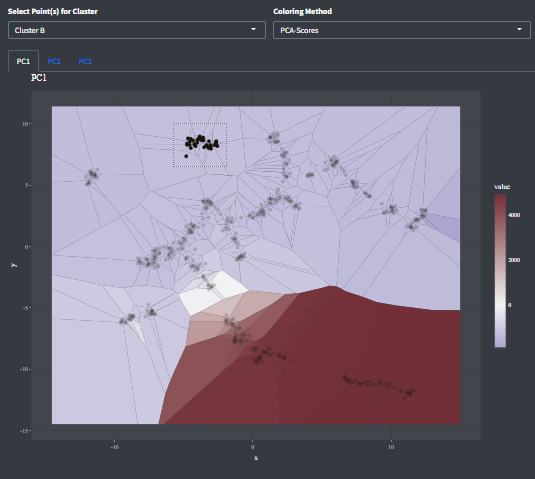}
  \captionof{figure}{Cluster B}
  \label{fig:viztool_clusterB}
  \end{subfigure}
  \caption{Cluster selection for computation of relative $T^2$-contributions. 
  \label{fig:viztool_clusters}
}
\end{figure}

\begin{figure}
\centering
  \begin{subfigure}[b]{0.45\textwidth}
\centering
	\includegraphics[height = 5.0cm]{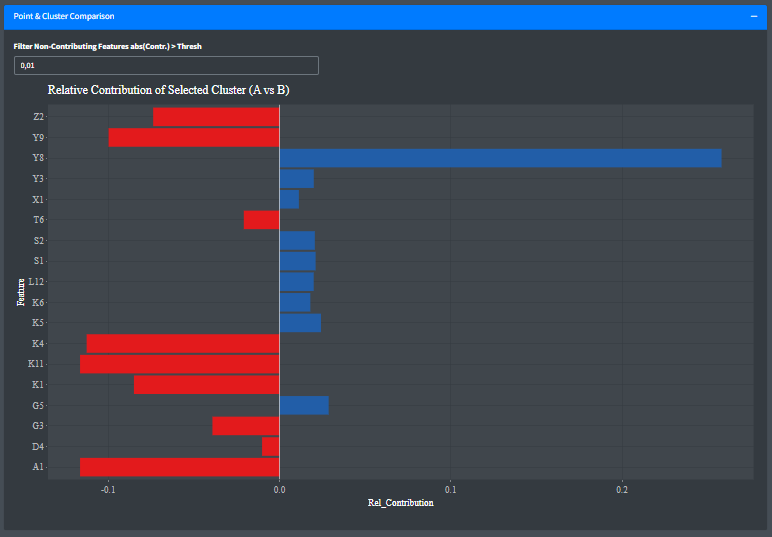}
  \label{fig:viztool_t2cont}
  \captionof{figure}{Relative $T^2$-contributions.}
  \end{subfigure}
  \hfill
  \begin{subfigure}[b]{0.45\textwidth}
  \centering
		\includegraphics[height = 5.5cm, page=1]{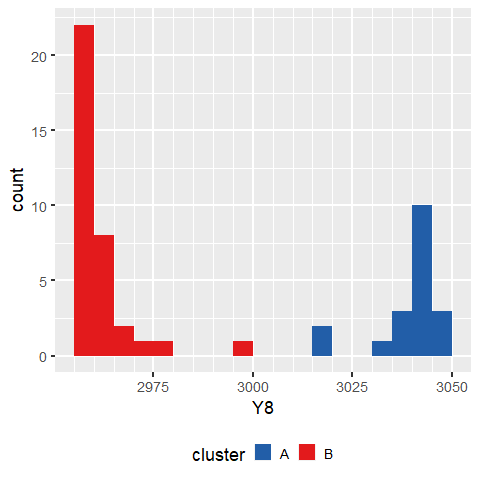}
  \captionof{figure}{Histogram of variable Y8}
  \label{fig:viztool_histogramY8}
  \end{subfigure}
  \caption{Comparison clusters A and B by means of relative Hotelling's $T^2$-contributions}
  \label{fig:viztool_relTcontr_hist}
\end{figure}

In some cases, exploring a non-linear embedding using a single PC at the time is too complex, which is especially true for data sets with a flat Eigenvalue distribution. We here propose relative Hotelling's $T^2$ contributions (see Eq. \eqref{eq:rel_Tcon}) in order to include information from all relevant PCs when deriving explanations for why pairs of data points or clusters differ from each other. Figure \ref{fig:viztool_relTcontr_hist} shows the relative $T^2$-contributions corresponding to the difference between the clusters A and B shown in Figure \ref{fig:viztool_clusters}. The results indicate that the two clusters differ mostly by the value of variable Y8, which does not become evident from the color-coded UMAP shown in Figure \ref{fig:viztool_Y8}. This is because the absolute difference between the two distributions is small compared to the total variance of the variable across the entire data set (Figure \ref{fig:viztool_histogramY8}). Altogether, this example underpins the strength of analyzing a UMAP-type embedding by means of relative $T^2$-contributions.

\section{Discussion}
\label{sec:discussion}
We proposed a novel workflow that employs the well-known Hotelling's $T^2$ statistic and $Q$-residuals from PCA to derive explanations for non-linear neighbor embedding such as UMAP. We integrated these diagnostic tools in a software prototype using the shiny web application framework for R and demonstrated how they can be used for exploratory analysis of a multivariate dataset with UMAP. In particular, we showed how our approach is capable of identifying discriminatory features between groups of data points that remain unnoticed when exploring NE using standard univariate (variable-by-variable) or multivariate (using PC-wise color-coding) approaches which underpins the strength of the approach.     

\section{Conclusion}
\label{sec:conclusion}
Non-linear dimension reduction techniques such as UMAP often provide valuable insights into high-dimensional data sets beyond those obtained by PCA. However, interpretation of the former is not as straight forward as with the latter and thus requires a downstream (confirmatory) data analysis pipeline. We found previously proposed methodology to "open the black box of neighbor embedding" limited in terms of both, user-friendliness and computational efficiency that would allow non-expert users to "make sense" out of NE in a consistent way. Our approach addresses both these shortcomings.

\section*{Acknowledgements}
Funding was provided by the Federal Ministry for Climate Action, Environment, Energy, Mobility, Innovation and Technology (BMK), the Federal Ministry for Digital and Economic Affairs (BMDW), and the Province of Upper Austria in the frame of the COMET - Competence Centers for Excellent Technologies programme managed by the Austrian Research Promotion Agency FFG and the COMET Center CHASE.

%% The Appendices part is started with the command \appendix;
%% appendix sections are then done as normal sections
%\appendix

%\section{Sample Appendix Section}
%\label{sec:sample:appendix}
%Lorem ipsum dolor sit amet, consectetur adipiscing elit, sed do eiusmod tempor section \ref{sec:sample1} incididunt ut labore et dolore magna aliqua. Ut enim ad minim veniam, quis nostrud exercitation ullamco laboris nisi ut aliquip ex ea commodo consequat. Duis aute irure dolor in reprehenderit in voluptate velit esse cillum dolore eu fugiat nulla pariatur. Excepteur sint occaecat cupidatat non proident, sunt in culpa qui officia deserunt mollit anim id est laborum.

%% If you have bibdatabase file and want bibtex to generate the
%% bibitems, please use
%%
 \bibliographystyle{elsarticle-num} 
 \bibliography{cas-refs}

\begin{thebibliography}{10}
\expandafter\ifx\csname url\endcsname\relax
  \def\url#1{\texttt{#1}}\fi
\expandafter\ifx\csname urlprefix\endcsname\relax\def\urlprefix{URL }\fi
\expandafter\ifx\csname href\endcsname\relax
  \def\href#1#2{#2} \def\path#1{#1}\fi

\bibitem{van2008visualizing}
L.~Van~der Maaten, G.~Hinton, Visualizing data using t-sne., Journal of machine
  learning research 9~(11) (2008).

\bibitem{li2017application}
W.~Li, J.~E. Cerise, Y.~Yang, H.~Han, Application of t-sne to human genetic
  data, Journal of bioinformatics and computational biology 15~(04) (2017)
  1750017.

\bibitem{chang2004super}
H.~Chang, D.-Y. Yeung, Y.~Xiong, Super-resolution through neighbor embedding,
  in: Proceedings of the 2004 IEEE Computer Society Conference on Computer
  Vision and Pattern Recognition, 2004. CVPR 2004., Vol.~1, IEEE, 2004, pp.
  I--I.

\bibitem{hajibabaee2021empirical}
P.~Hajibabaee, F.~Pourkamali-Anaraki, M.~A. Hariri-Ardebili, An empirical
  evaluation of the t-sne algorithm for data visualization in structural
  engineering, arXiv preprint arXiv:2109.08795 (2021).

\bibitem{mcinnes2018umap}
L.~McInnes, J.~Healy, J.~Melville, Umap: Uniform manifold approximation and
  projection for dimension reduction, arXiv preprint arXiv:1802.03426 (2018).

\bibitem{smilkov2016embedding}
D.~Smilkov, N.~Thorat, C.~Nicholson, E.~Reif, F.~B. Vi{\'e}gas, M.~Wattenberg,
  Embedding projector: Interactive visualization and interpretation of
  embeddings, arXiv preprint arXiv:1611.05469 (2016).

\bibitem{chatzimparmpas2020t}
A.~Chatzimparmpas, R.~M. Martins, A.~Kerren, t-visne: Interactive assessment
  and interpretation of t-sne projections, IEEE transactions on visualization
  and computer graphics 26~(8) (2020) 2696--2714.

\bibitem{bibal2020explaining}
A.~Bibal, V.~M. Vu, G.~Nanfack, B.~Fr{\'e}nay, Explaining t-sne embeddings
  locally by adapting lime., in: ESANN, 2020, pp. 393--398.

\bibitem{andries2022dual}
E.~Andries, R.~Nikzad-Langerodi, Dual-constrained and primal-constrained
  principal component analysis, Journal of Chemometrics (2022) e3403.

\bibitem{wang2020understanding}
Y.~Wang, H.~Huang, C.~Rudin, Y.~Shaposhnik, Understanding how dimension
  reduction tools work: an empirical approach to deciphering t-sne, umap,
  trimap, and pacmap for data visualization, arXiv preprint arXiv:2012.04456
  (2020).

\bibitem{jolliffe2005principal}
I.~Jolliffe, Principal component analysis, Encyclopedia of statistics in
  behavioral science (2005).

\bibitem{wold1987principal}
S.~Wold, K.~Esbensen, P.~Geladi, Principal component analysis, Chemometrics and
  intelligent laboratory systems 2~(1-3) (1987) 37--52.

\bibitem{ferrer2007multivariate}
A.~Ferrer, Multivariate statistical process control based on principal
  component analysis (mspc-pca): Some reflections and a case study in an
  autobody assembly process, Quality Engineering 19~(4) (2007) 311--325.

\bibitem{wold1977simca}
S.~Wold, M.~Sj{\"o}str{\"o}m, Simca: a method for analyzing chemical data in
  terms of similarity and analogy, ACS Publications, 1977.

\bibitem{nikzad2020domain}
R.~Nikzad-Langerodi, W.~Zellinger, S.~Saminger-Platz, B.~A. Moser, Domain
  adaptation for regression under beer--lambert’s law, Knowledge-Based
  Systems 210 (2020) 106447.

\bibitem{chang2017shiny}
W.~Chang, J.~Cheng, J.~Allaire, Y.~Xie, J.~McPherson, et~al., Shiny: web
  application framework for r, R package version 1~(5) (2017) 2017.

\end{thebibliography}

%% else use the following coding to input the bibitems directly in the
%% TeX file.

% \begin{thebibliography}{00}

% %% \bibitem{label}
% %% Text of bibliographic item

% \bibitem{}

% \end{thebibliography}
\end{document}